\documentclass[11pt]{article}

\usepackage[]{acl}

\usepackage{times}
\usepackage{latexsym}

\usepackage[T1]{fontenc}
\usepackage[utf8]{inputenc}

\usepackage{microtype}

\usepackage{inconsolata}
\usepackage{inconsolata}

\usepackage{blindtext}
\usepackage{booktabs}
\usepackage{lscape} 
\usepackage{graphicx}
\usepackage{subcaption}
\usepackage{xspace}
\usepackage{arydshln}
\usepackage{bbm}
\usepackage[normalem]{ulem}
\usepackage{multirow}
\usepackage{array,tabularx}
\usepackage{todonotes}
\usepackage{amssymb}
\usepackage[acronym]{glossaries}
\usepackage{makecell}
\usepackage{textcomp}
\usepackage{rotating,booktabs}
\usepackage{cleveref}
\usepackage{subcaption}
\usepackage{enumitem}
\usepackage{soul}
\usepackage{float}

\usepackage{booktabs}
\usepackage{multirow}
\usepackage{hyperref}
\usepackage{cleveref}
\usepackage{ragged2e}

\usepackage{xcolor}
\usepackage{tcolorbox}
\usepackage{geometry}

\usepackage{graphicx}
\graphicspath{{figures/}}

\usepackage{amsmath}
\usepackage{amssymb}
\usepackage{booktabs}

\title{Which Objectives Need a Dial? Predicting Objective Conflict and Covering Trade-offs in Steerable Pluralistic Alignment}

\author{David Tsoi$^{1}$ \and Esra Dönmez$^{2,3}$\\
    $^1$Department of Computer Science, University of Stuttgart\\
    $^2$Institute for Natural Language Processing, University of Stuttgart\\
    $^3$Interchange Forum for Reflecting on Intelligent Systems, University of Stuttgart\\
    \normalsize{\texttt{\{david-alexander.tsoi,esra.doenmez\}@ims.uni-stuttgart.de}}}

\begin{document}
\maketitle

\begin{abstract}
People hold diverse, sometimes conflicting values, so no single aligned model can satisfy everyone. Pluralistic alignment therefore calls for steerable models that can balance competing objectives differently. Multi-Objective Direct Preference Optimization (MODPO) does this by using an objective weight to span a continuum of trade-offs. We study two questions: when can one model improve two objectives simultaneously, and how can many trade-offs be covered without training a separate model for each? Across seven objective pairs from HelpSteer and UltraFeedback, two pre-training measurements predict whether objectives align or conflict for human-annotated data, but not for AI-annotated data, where response length and repetition confound reward-model scores. For broader trade-off coverage, selecting the nearest trained model and merging model parameters both help, but neither consistently matches direct training. These findings yield practical guidance for building steerable models that serve diverse preferences.
\end{abstract}

\section{Introduction}
\label{sec:introduction}

People do not share one set of values. They disagree about what makes a response good, and the qualities they care about trade off against one another: a longer answer can be more helpful but less concise, and a maximally cautious answer can be less useful. When we align a model, these qualities become \emph{objectives} we optimize, and different stakeholders, holding different values, weight the same objectives differently over the same preferences. Aligning to a single averaged preference hides this diversity and picks winners among objectives that people would weight differently \citep{bakker2022finetuning,plank-2022-problem}. \emph{Pluralistic alignment} asks instead how one system can serve a plurality of stakeholders \citep{sorensen2024roadmap}.

One practical route is \emph{steerable} pluralism: rather than fixing one compromise, build a model that can be steered to weight competing objectives differently \citep{sorensen2024roadmap}. Multi-Objective Direct Preference Optimization \citep[MODPO;][]{zhou-etal-2024-beyond} is an efficient way to build such a model. It extends Direct Preference Optimization \citep[DPO;][]{rafailov2023dpo} with an \emph{objective weight} that steers the balance between two objectives, such as helpfulness, honesty, or safety. Training at a range of weights yields a family of models spanning a continuum of trade-offs, so a stakeholder can pick the compromise their context calls for.

Two questions decide whether MODPO can be used this way in practice, and neither is answered by reporting scores at a single weight.

\paragraph{RQ1: When can one model satisfy two objectives at once?} Some objectives reinforce each other and some conflict. If MODPO can improve two objectives together, a practitioner should train one model that serves both; if they conflict, they must instead steer along a trade-off and choose a point on it. We ask whether this can be predicted \emph{before} training from properties of the preference data and the base model. We study two cheap measurements: the
\textbf{preference-pair conflict rate}, how often the two objectives disagree on which response is better \citep{li-etal-2025-self-improvement}, a direct measure of that disagreement in the data \citep{plank-2022-problem}; and the \textbf{SFT reference-model score correlation}, whether the SFT reference model's better answers for one objective are also better for the other.

\paragraph{RQ2: How can we cover many trade-offs without training a model for each?} Serving a continuum of stakeholder preferences should not require training one model per compromise. Given a family of MODPO models already trained at a few weights, we ask how closely two training-free reuse methods match a model trained directly at a new \emph{target} weight: \textbf{nearest-model selection}, using the trained model whose weight is closest to the target, and \textbf{parameter merging}, blending the LoRA adapters of the two neighbouring trained models \citep{jang2023personalizedsoups,rame2023rewardedsoups}.

\paragraph{Contributions.} (i) We frame MODPO as a tool for steerable pluralistic alignment and ask, concretely, when it can satisfy two objectives jointly and when it must trade them off. (ii) We test two pre training measurements of objective compatibility across seven objective pairs. They predict the outcome well for the human-annotated HelpSteer data but fail for the AI-annotated UltraFeedback data. (iii) We trace that failure to a \emph{sociotechnical} cause: the apparent conflict is strongly confounded by response length and repetition in the reward-model scores, a warning for how objective behavior is evaluated. (iv) We show that nearest-model selection and parameter merging can cover the trade-off continuum without retraining, but neither reliably matches direct training. We distill each finding into a practical recipe.

\section{Related Work}
\label{sec:related}

\paragraph{Pluralistic alignment.} A growing line of work argues that alignment should reflect diverse and potentially conflicting human values rather than collapse them into a single averaged preference \citep{sorensen2024roadmap,conitzer2024socialchoice,kirk2024prism}. This concern is reinforced by evidence that reward-model quality varies across socially consequential domains and that reward models can systematically prefer socially undesirable responses \citep{ghazaryan2026misalignedrewardsociallyundesirable}. \citet{sorensen2024roadmap} distinguish several forms of pluralism; we focus on the \emph{steerable} setting, in which a single model can be adjusted to weight its objectives differently for different stakeholders. Existing approaches include training on population-level agreement \citep{bakker2022finetuning}, optimizing max-min objectives over diverse preferences \citep{chakraborty2024maxmin}, and coordinating multiple LLMs \citep{feng-etal-2024-modular}. We study MODPO as a concrete and efficient approach to steerable pluralism, asking when a single model can satisfy two objectives jointly and when those objectives instead require a trade-off.

\paragraph{Steering multiple objectives.} DPO \citep{rafailov2023dpo} aligns a model to preferences without a separate reward model or reinforcement learning. Several methods extend alignment to several objectives with a controllable balance: MODPO folds the balance into the DPO loss \citep{zhou-etal-2024-beyond}; directional preference alignment attaches a preference direction to a multi-objective reward \citep{wang-etal-2024-arithmetic}; Rewards-in-Context conditions on the balance in the prompt \citep{yang2024ric}; and controlled or multi-objective decoding sets it at generation time \citep{mudgal2024controlled,shi2024mod}. These works report scores at chosen weights; we instead ask whether the \emph{outcome} at each weight can be predicted before training.

\paragraph{Disagreement in preference data.} Human annotators disagree, and this variation is signal, not just noise \citep{plank-2022-problem,bakker2022finetuning}. For multiple objectives, \citet{li-etal-2025-self-improvement} name \emph{preference conflict} (the two objectives prefer different responses for the same pair), show it is common, and reduce it by rewriting responses; \citet{xu-etal-2026-understanding} study a related reward consistency. We do not try to remove conflict. We measure it, and whether it and a base-model correlation predict which objectives MODPO can jointly satisfy.

\paragraph{Reusing and merging models.} Weight-space methods combine trained models without retraining: Model Soups \citep{wortsman2022modelsoups}, Task Arithmetic \citep{ilharco2023taskarithmetic}, and AdapterSoup \citep{chronopoulou-etal-2023-adaptersoup}. For alignment specifically, Rewarded Soups \citep{rame2023rewardedsoups} and Personalized Soups \citep{jang2023personalizedsoups} merge models trained on different rewards to reach new preference balances. Our parameter-merging method blends the LoRA adapters \citep{hu2022lora} of the two trained MODPO models nearest a target weight, and we compare it against nearest-model selection and direct training.

\paragraph{Evaluating with reward models.} Reward models enable scalable, consistent scoring across objectives, and multi-objective models such as ArmoRM \citep{wang-etal-2024-arithmetic} let us evaluate both objectives on the same generations. Because scores can be influenced by proxy optimization, response length, category, and repetition \citep{gao2023overoptimization,singhal2024length,dubois2024alpacaeval,zheng2023llmjudge,lambert-etal-2025-rewardbench,welleck2020unlikelihood}, we report length and repetition alongside reward scores and re-decode with repetition control. We show that these checks are important for interpreting whether two objectives are genuinely compatible.
\section{Background}
\label{sec:background}

We align to two objectives at a time and always fix the first as helpfulness.

\paragraph{Preference data and conflict.} A \emph{response pair} $(x, y_i, y_j)$ is a prompt $x$ with two responses. Each objective $o$ scores responses; the higher-scoring response is preferred, giving a \emph{preference pair} $(x, y_w, y_l)$ with preferred $y_w$ and dispreferred $y_l$ (ties are dropped). One response pair thus yields one preference pair \emph{per} objective, and the two need not agree. When two objectives prefer different responses for the same pair, they are in \emph{conflict} \citep{li-etal-2025-self-improvement}. Over a dataset, the fraction of jointly labelled pairs that conflict is the \emph{preference-pair conflict rate}. It is a property of the data, measures how much the two objectives disagree, and is computed before any training.

\paragraph{SFT reference model.} Preference optimization does not start from the pretrained model directly: it starts from a \emph{reference model} $\pi_{\text{sft}}$, obtained by \emph{supervised fine-tuning} (SFT), i.e.\ plain next-token likelihood training on demonstration responses (here, following \citet{zhou-etal-2024-beyond}, the preferred responses of the helpfulness preference pairs). SFT uses no preference signal: it never sees the dispreferred response and never sees the second objective. This one model plays three roles in what follows. It initializes every MODPO run; it is the fixed anchor MODPO keeps the trained policy close to (via $\beta$ below); and it is the model whose generations we score before training in \S\ref{sec:measurements}. Because all trained models share it as a starting point, it is also the baseline that their scores are read against: a model that already answers in the dataset's domain and format, but has not yet been optimized for either objective of a pair.

\paragraph{DPO.} DPO treats a model as a policy $\pi_\theta(y\mid x)$ and trains it against the reference policy $\pi_{\text{sft}}$, which stays frozen \citep{rafailov2023dpo}. Writing
$\Delta_\theta(x,y) = \log\pi_\theta(y\mid x) - \log\pi_{\text{sft}}(y\mid x)$ and the preference margin $d_\theta = \Delta_\theta(x,y_w) - \Delta_\theta(x,y_l)$, the loss is
\begin{equation}
\label{eq:dpo}
\mathcal{L}_{\text{DPO}}
= -\,\mathbb{E}_{(x,y_w,y_l)}\big[\log\sigma(\beta\,d_\theta)\big],
\end{equation}
where $\sigma$ is the logistic function and $\beta$ limits how far the policy drifts from the reference. DPO optimizes one objective.

\paragraph{MODPO.} MODPO trains one policy for a weighted mix of two objectives \citep{zhou-etal-2024-beyond}. An \emph{objective weight} $w \in (0,1]$ sets the balance: the preference pairs come from the first objective (helpfulness), while the second enters through a margin $m_\phi = r_\phi(x,y_w) - r_\phi(x,y_l)$ from a reward model $r_\phi$ trained on it. The DPO logit becomes
\begin{equation}
\label{eq:modpo}
z = \tfrac{1}{w}\big(\beta\,d_{\theta_w} - (1-w)\,m_\phi\big),
\end{equation}
with loss $-\mathbb{E}[\log\sigma(z)]$. Each weight $w$ gives a separate model at a \emph{known} point on the trade-off, so a set of weights is a steerable family. Flipping the sign of the margin reverses which direction of the second objective is preferred (e.g.\ shorter vs.\ longer responses).

\paragraph{LoRA.} We train with Low-Rank Adaptation \citep{hu2022lora}: the pretrained weights $W_0$ are frozen and a low-rank update $\Delta W = BA$ is learned, applied as $h = W_0 u + (\alpha/r)\,BA\,u$. A trained model is then the base model plus a small adapter, so storing many objective-weight variants is cheap and their adapters can be merged (\S\ref{sec:method}).

\section{Method}
\label{sec:method}

We run two studies\footnote{\url{https://github.com/esradonmez/modpo-objective-trade-off}}. The \emph{trade-off study} (RQ1) asks whether pre-training measurements predict which MODPO models improve two objectives together and which trade them off. The \emph{coverage study} (RQ2) asks how well trained models can be reused to reach new trade-offs without retraining. We report two kinds of score and keep them distinct: \emph{dataset scores}, which ship with the data and are used only to build preference pairs, and \emph{reward-model scores}, which the evaluator assigns to generated responses and which we compare.

\subsection{Measuring objective compatibility before training}
\label{sec:measurements}

For each objective pair we compute two cheap measurements before any MODPO training.

\paragraph{Preference-pair conflict rate.} Over the training data, we take the response pairs that yield a preference pair for \emph{both} objectives and measure how often the two prefer different responses (\S\ref{sec:background}). This is a direct count of how often the objectives disagree in the data. A high rate means they pull toward opposite responses, which is what produces the conflicting gradients that can force a trade-off.

\paragraph{SFT reference-model score correlation.} We generate one response per evaluation prompt from the reference model, score it for both objectives, and take the Pearson correlation of the two score vectors. The correlation is high when the reference model's higher-scoring responses for one objective are also its higher-scoring responses for the other. We compute it on \emph{generated} responses with the same prompts and evaluator later used for the MODPO models, so it is comparable to their scores. A correlation of the raw dataset scores would not be, since those come from a different annotator (humans for HelpSteer, GPT-4 for UltraFeedback).

A high correlation and a low conflict rate both point to objectives that move together, so we expect them to agree. The interesting cases are where they do not, and testing those is crucial for understanding alignment objectives.

\subsection{Probing the conflict rate}
\label{sec:selection}

To test the conflict rate more directly, we split the jointly labelled pairs into a \emph{non-conflict} set (rate $0$) and a \emph{conflict} set (rate $1$), keep the helpfulness preference pairs fixed, and retrain only the second-objective margin reward model on each subset. Because the subsets also differ in size and composition, we read these results as descriptive, not causal.

\subsection{Covering the trade-off continuum at inference time}
\label{sec:inference}

A \emph{target} trade-off is an objective weight $w^\star$ chosen after training. We compare two training-free reuse methods against a model trained directly at $w^\star$.

\paragraph{Nearest-model selection.} From a set $G$ of trained weights, pick $w_g = \arg\min_{w\in G}|w-w^\star|$ (ties broken toward the lower weight) and generate with that model. We compare a \emph{fine} set (more weights) and a \emph{coarse} set (fewer).

\paragraph{Parameter merging.} Using the fine set, take the closest lower and upper trained weights $w_\ell < w^\star < w_u$, set $\lambda = (w^\star - w_\ell)/(w_u - w_\ell)$, and merge their LoRA adapters tensor by tensor, $\Theta_{\text{merge}} = (1-\lambda)\,\Theta_\ell + \lambda\,\Theta_u$ \citep{jang2023personalizedsoups,rame2023rewardedsoups}. No new training is done. Merging is an interpolation heuristic, not an optimization procedure for $w^\star$: the MODPO objective at $w^\star$ is never evaluated, so nothing guarantees the merged adapter behaves like one trained there. How closely it does is what we measure.

\section{Experimental Setup}
\label{sec:experiments}

\paragraph{Datasets.} We use two datasets that score more than two objectives so that several objective pairs share one dataset. HelpSteer \citep{wang-etal-2024-helpsteer} provides \emph{human}-assigned scores ($0$--$4$) for helpfulness, correctness, coherence, complexity, and verbosity over $10{,}459$ prompts. UltraFeedback \citep{cui2024ultrafeedback} provides \emph{GPT-4}-assigned scores ($1$--$5$) for helpfulness, honesty, instruction-following, and truthfulness over $63{,}967$ prompts. The coverage study uses PKU-SafeRLHF-10K \citep{ji2023beavertails,dai2024saferlhf} with the helpfulness/harmlessness pair (an objective pair where the trade-off is real and most consequential) reproducing the MODPO safety setup so our numbers can be checked against \citet{zhou-etal-2024-beyond}.

\paragraph{Objectives.} We fix \emph{helpfulness} as the first objective in every pair and choose the second objectives to span ones that plausibly reinforce helpfulness and ones that plausibly compete with it: from HelpSteer, \emph{correctness} and \emph{coherence} should reinforce it, \emph{complexity} is more independent, and \emph{verbosity} can compete with it (a concise answer may still be helpful); from UltraFeedback, \emph{honesty}, \emph{instruction-following}, and \emph{truthfulness} are broadly aligned with helpfulness but need not coincide with it response by response. This gives four HelpSteer pairs and three UltraFeedback pairs; for HelpSteer verbosity we train once toward higher and once toward lower verbosity (positive / negative margin sign), for eight settings in all. The spread lets us test the measurements on pairs expected to agree and pairs expected to conflict, rather than on a single case.

\paragraph{Models and training.} We use \textsc{Llama-2-7b} \citep{touvron2023llama2} for HelpSteer and \textsc{Alpaca-7b} \citep{taori2023alpaca} for UltraFeedback and PKU-SafeRLHF, all trained with LoRA. Following \citet{zhou-etal-2024-beyond}, each setting trains a margin reward model on the second objective and one MODPO model per objective weight. In the trade-off study we additionally train an SFT reference model on the preferred responses of the helpfulness pairs and use weights $w\in\{0.1,0.2,0.4,0.6,0.8,(0.9)\}$. The coverage study reuses the \textsc{Alpaca-7b} checkpoint as its reference model without further SFT, again following \citet{zhou-etal-2024-beyond}, and trains the \emph{fine} weight set $\{0.0,0.2,0.4,0.6,0.8,1.0\}$, the \emph{coarse} set $\{0.0,0.5,1.0\}$, and direct models at the target weights $w^\star\in\{0.1,0.3,0.5,0.7,0.9\}$; at $w{=}0.0$ all weight is on harmlessness, so the DPO margin adapter is loaded instead of a MODPO adapter. We report the \emph{2048-token} training-length limit as primary (it retains almost all pairs) and contrast a \emph{512-token} limit that retains far fewer. Training uses learning rate $10^{-4}$, $\beta=0.1$, AdamW with a cosine schedule, and a single seed per model; full hyperparameters are in Appendix~\ref{sec:appendix}.

\paragraph{Evaluation.} We generate one greedy response per prompt on $400$ held-out prompts (trade-off study) or $700$ response pairs (coverage study). Trade-off responses are scored per objective with ArmoRM \citep{wang-etal-2024-interpretable}; PKU responses are scored with the Beaver reward
(helpfulness, higher better) and cost (harmlessness, lower better) models \citep{dai2024saferlhf}. We report mean scores, mean response length in words, and the share of responses with a repeated $4$-gram, with $95\%$ paired bootstrap confidence intervals \citep{efron1993bootstrap}. Because greedy decoding can produce long, repetitive text, we additionally re-decode the 2048-token models with a repetition penalty and $4$-gram blocking to test whether score changes track length and repetition.

\begin{figure}[t]
  \centering
  \includegraphics[width=\columnwidth]{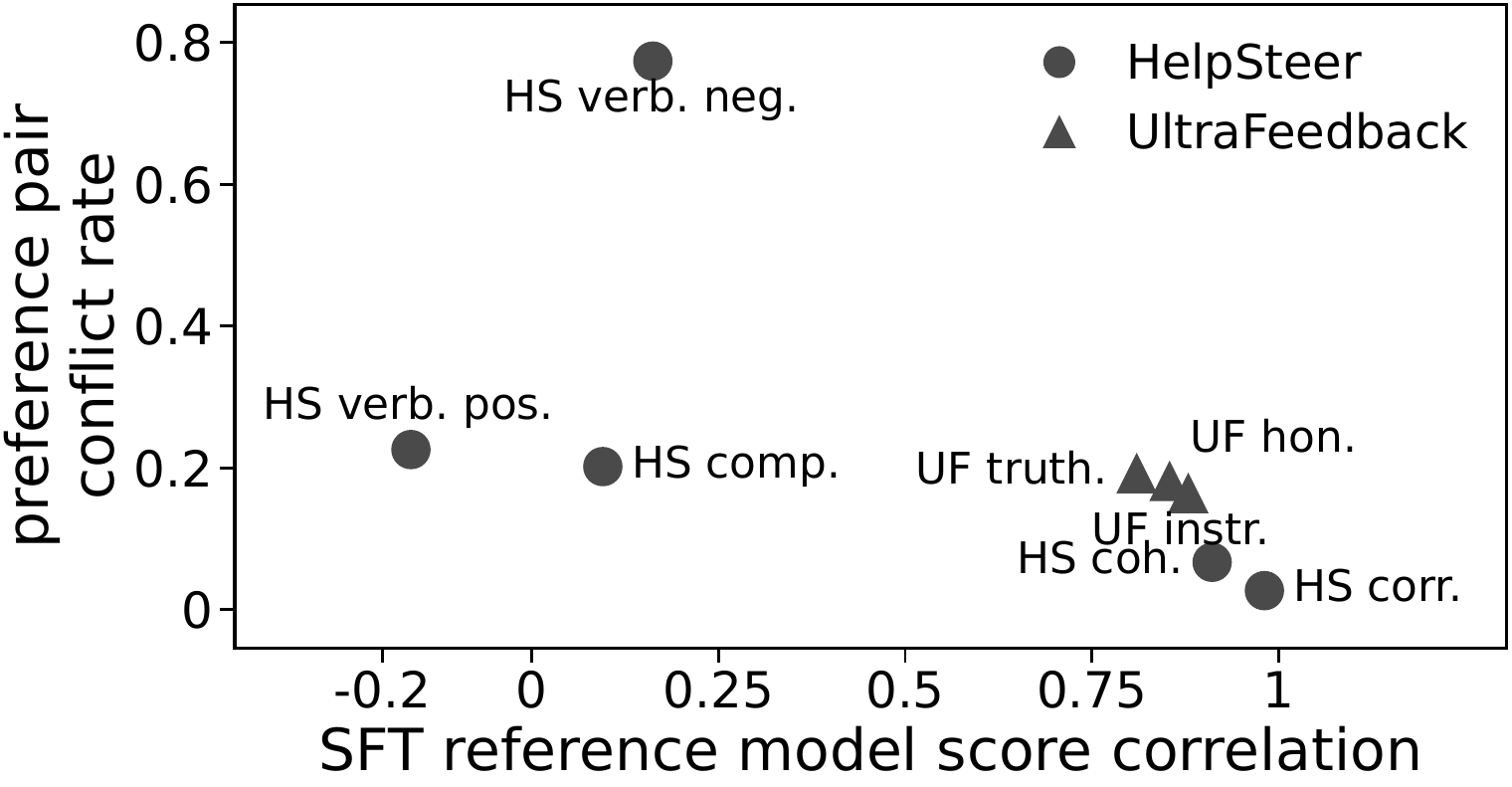}
  \caption{The two pre-training measurements for all seven objective pairs (2048-token limit; both verbosity margin signs are plotted, giving eight points). Objectives that move together sit toward the bottom right (high correlation, low conflict); HelpSteer correctness and coherence are the clearest such cases, while HelpSteer verbosity toward concision (\emph{HS verb.\ neg.}) has by far the highest conflict rate. The two measurements agree for the three UltraFeedback pairs too: high correlation and a conflict rate below every HelpSteer pair but correctness and coherence. Verbosity is where the two measurements part company; UltraFeedback is where the \emph{results} will.}
  \label{fig:scatter}
\end{figure}

\begin{table}[t]
\centering
\small
\setlength{\tabcolsep}{4pt}
\begin{tabular}{@{}lccc@{}}
\toprule
\textbf{Objective pair} & \textbf{Corr.} & \textbf{Conflict} & \textbf{Both$\uparrow$?} \\
\midrule
\multicolumn{4}{@{}l}{\emph{HelpSteer}} \\
\quad correctness            & $0.981$  & $0.027$ & all $w$ \\
\quad coherence              & $0.911$  & $0.067$ & all $w$ \\
\quad complexity             & $0.095$  & $0.202$ & 2 of 5 $w$ \\
\quad verbosity ($+$)        & $-0.162$ & $0.226$ & $w{=}0.9$ \\
\quad verbosity ($-$)        & $0.162$  & $0.774$ & none$^\dagger$ \\
\midrule
\multicolumn{4}{@{}l}{\emph{UltraFeedback}} \\
\quad honesty                & $0.854$  & $0.181$ & none \\
\quad instruction-following  & $0.879$  & $0.164$ & none \\
\quad truthfulness           & $0.810$  & $0.192$ & none \\
\bottomrule
\end{tabular}
\caption{Pre-training measurements (SFT reference-model score correlation and preference-pair conflict rate, 2048-token limit) and at which objective weights \emph{both} objectives improve over the reference under greedy decoding at the same limit. $^\dagger$verbosity($-$) improves both only at $w{=}0.4$ under the 512-token limit. Appendix~\ref{sec:app-counts} gives the 512-token measurements and Appendix~\ref{sec:app-scores} the underlying scores. HelpSteer follows the measurements; the high-correlation UltraFeedback pairs do not.}
\label{tab:measurements}
\end{table}

\section{Results}
\label{sec:results}

\subsection{When can one model satisfy two objectives? (RQ1)}
\label{sec:results-tradeoff}

\begin{figure*}[t]
  \centering
  \includegraphics[width=0.48\textwidth]{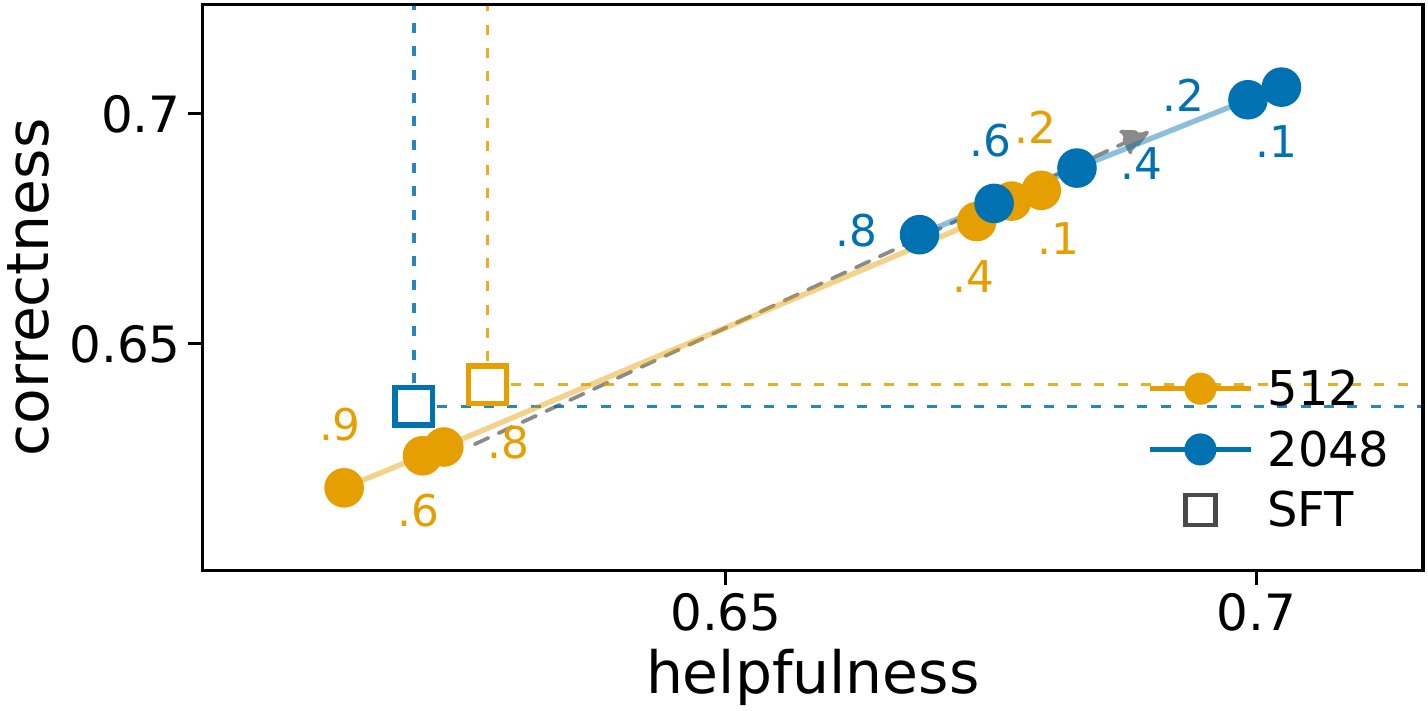}
  \hfill
  \includegraphics[width=0.48\textwidth]{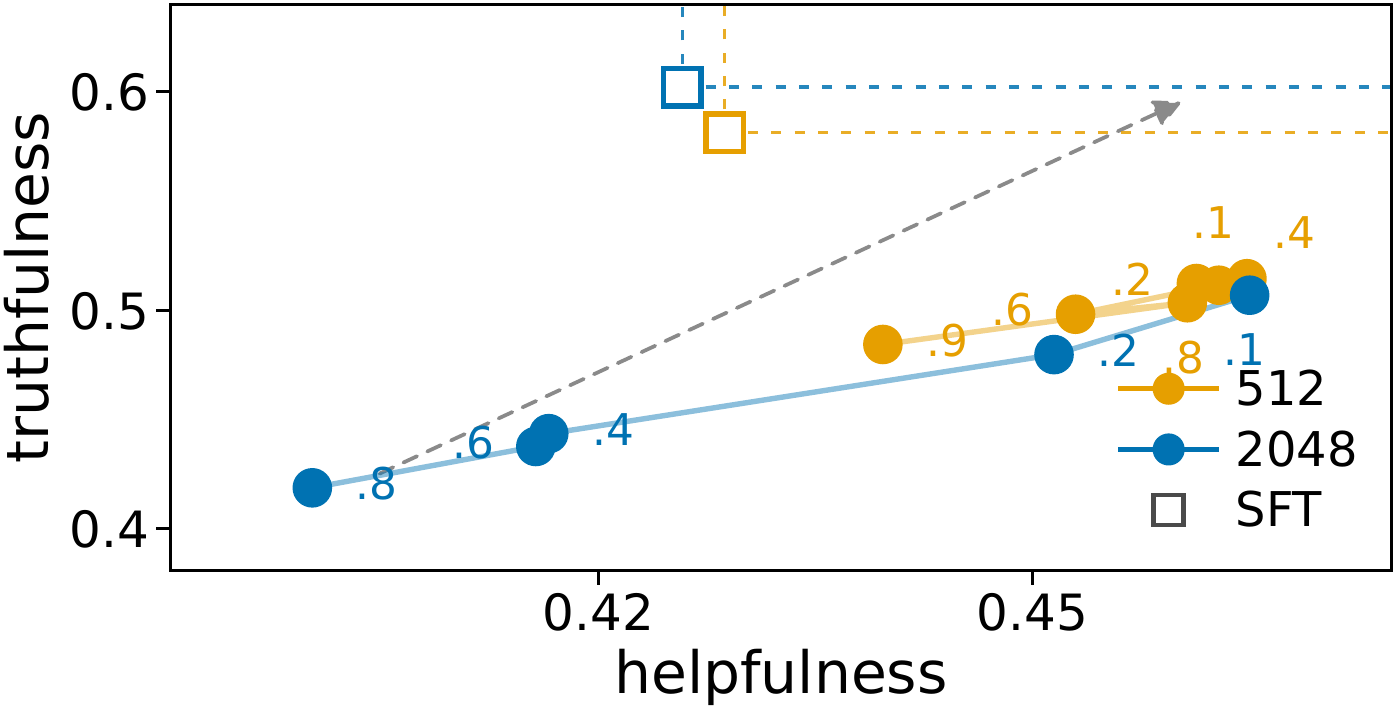}
  \caption{MODPO scores (ArmoRM mean over 400 prompts) for an objective pair that MODPO can satisfy jointly and one it cannot, at two training-length limits (512, 2048). Empty squares mark the reference model; labels are objective weights $w$. \textbf{Left:} HelpSteer correctness. Under the 2048-token limit every MODPO model improves \emph{both} objectives relative to its reference model; the 512-token condition shows weaker and less consistent gains. \textbf{Right:} UltraFeedback truthfulness. Despite a high pre-training correlation, every model scores \emph{lower} on truthfulness than the reference. The gap between the 512 and 2048 clouds shows how much the training-length limit moves the scores.}
  \label{fig:pairs}
\end{figure*}

\paragraph{The measurements work for HelpSteer.} \Cref{tab:measurements} and Figure~\ref{fig:scatter} display the two pre-training measurements; Figure~\ref{fig:pairs} shows the resulting MODPO scores for one pair from each dataset. For HelpSteer the measurements match both intuition and each other. \emph{Correctness} and \emph{coherence} have very high correlations ($0.98$, $0.91$) and low conflict rates ($0.03$, $0.07$). Under the 2048-token limit, MODPO improves \emph{both} objectives at every weight (the ``all $w$'' entries in \Cref{tab:measurements}): in Figure~\ref{fig:pairs} (left) every 2048-token model sits up and to the right of the reference, with correctness rising from $0.636$ to $0.706$ at $w{=}0.1$ as helpfulness rises from $0.621$ to $0.702$; the 512-token condition shows weaker and less consistent gains. \emph{Complexity} has a weak correlation ($0.095$) and moderate conflict ($0.202$): it improves at every weight but helpfulness only at two, a partial match. \emph{Verbosity toward concision} has the highest conflict rate ($0.774$) and shows a clean trade-off: low weights favor concision, high weights favor helpfulness, exactly as the conflict rate predicts.

\paragraph{The measurements fail for UltraFeedback.} The three UltraFeedback pairs have high correlations ($0.81$--$0.88$) and only moderate conflict, yet every model \emph{degrades} performance on the second objective at every weight and never improves both. Figure~\ref{fig:pairs} (right) highlights the contrast: every UltraFeedback model scores \emph{below} the reference on truthfulness. A high pre-training correlation did not deliver the joint improvement it did for HelpSteer.

\paragraph{The apparent conflict is strongly confounded by length and repetition.} Almost all MODPO models generate much longer and more repetitive responses than the reference (up to ${\sim}970$ words with a repeated-$4$-gram share above $0.95$ for UltraFeedback). When we re-decode the same 2048-token models with repetition control, mean length roughly halves and the repeated-$4$-gram share drops by $60$--$70$ points (\Cref{tab:repcontrol}; $38.7$--$81.6$ across individual pairs, Appendix~\ref{sec:app-repcontrol}). For HelpSteer this barely moves the scores. For UltraFeedback both objective scores rise by $+0.10$ to $+0.15$, and the resulting scores are substantially more consistent with the high reference-model correlations. This does not isolate the cause: re-decoding changes length, repetition, wording, and content together, and the same reward model scores both conditions, so we cannot separate a reward-model bias against repetition from genuinely better responses. What it does show is that the apparent UltraFeedback conflict is strongly confounded by generation length and repetition, and was driven at least partly by decoding behavior rather than by objective incompatibility alone; an independent evaluator would be needed to settle this. A plausible contributing factor is that UltraFeedback's GPT-4 scores track length more strongly than HelpSteer's human scores, and that HelpSteer separates verbosity into its own objective. The training-length limit adds to this: keeping more pairs shortens HelpSteer generations but lengthens UltraFeedback ones, which is why the same limit helps one dataset and hurts the
other (Figure~\ref{fig:pairs}).

\begin{table}[t]
\centering
\small
\setlength{\tabcolsep}{3pt}
\begin{tabular}{@{}lcccc@{}}
\toprule
& $\Delta$help & $\Delta$obj$_2$ & $\Delta$words & $\Delta$rep \\
\midrule
HelpSteer (mean)     & $-0.001$ & $-0.008$ & $-48.8\%$ & $-69.7$ \\
UltraFeedback (mean) & $+0.137$ & $+0.125$ & $-58.9\%$ & $-62.7$ \\
\bottomrule
\end{tabular}
\caption{Change in scores, response length, and repeated-$4$-gram share (percentage points) when the 2048-token models are re-decoded with repetition control. HelpSteer is unmoved; UltraFeedback rises sharply, showing its greedy-decoding scores were confounded by length and repetition.}
\label{tab:repcontrol}
\end{table}

\begin{figure}[t]
  \centering
  \includegraphics[width=0.92\columnwidth]{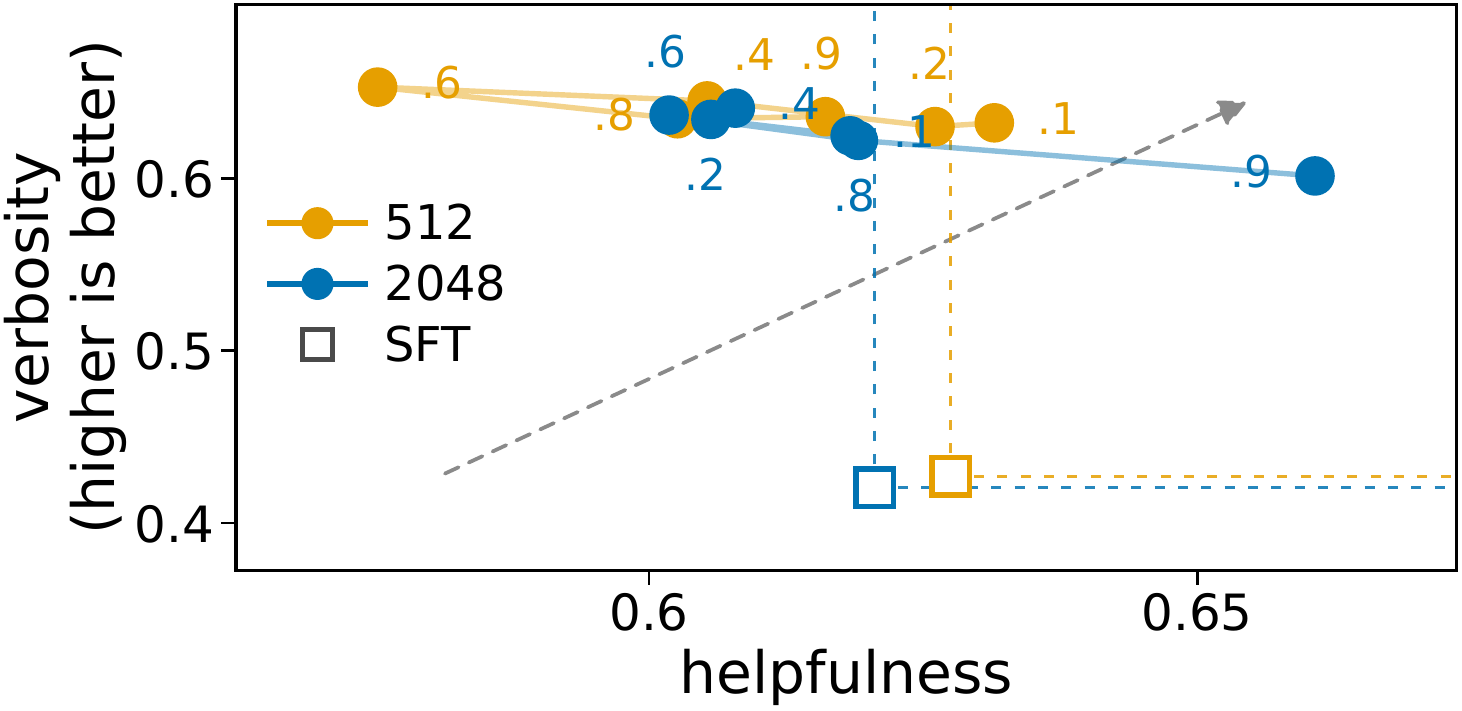}
  \caption{HelpSteer helpfulness--verbosity under the positive margin sign (steering toward higher verbosity). MODPO raises verbosity at every weight but improves helpfulness only at $w{=}0.9$, where helpfulness receives the greatest weight and verbosity the least. Flipping the sign steers toward concision and yields the trade-off in the text.}
  \label{fig:margin}
\end{figure}

\paragraph{The margin sign steers which direction of an objective wins.} By Eq.~\eqref{eq:modpo}, flipping the sign of the margin reverses which direction of the second objective is preferred. For verbosity this is the difference between steering toward longer and toward shorter responses. This is the explicit, per-objective control that steerable pluralism calls for. Figure~\ref{fig:margin} shows the positive sign: MODPO reliably raises verbosity but helps helpfulness at a single weight ($w{=}0.9$, the one setting where this pair
improves both), while the negative sign gives the clean trade-off above. The negative sign is also the pair with the highest conflict rate, so its trade-off is what the measurement predicts. Direction is therefore a second control knob, orthogonal to the weight.

\paragraph{Splitting on conflict changes the outcome but does not decide it.} Training the margin reward model only on conflicting pairs shifts the results but not all the way to a trade-off. For UltraFeedback instruction-following (the lowest conflict rate of the three UltraFeedback pairs; 512-token limit), the conflict set comes \emph{close} to a trade-off but stops short, because the pair still has a high correlation. HelpSteer verbosity toward concision points the same way from the other side: the split is computed in the dataset's score direction, so under the negative margin sign the ``conflict'' set is the one whose pairs \emph{agree} with helpfulness, and it duly improves helpfulness at every weight while worsening concision at the higher ones. Conflict alone is thus not enough; the correlation still matters, in line with \citet{li-etal-2025-self-improvement}. Full scores are in Appendix~\ref{sec:app-selection}.

\subsection{Covering the continuum without retraining (RQ2)}
\label{sec:results-inference}

\begin{figure}[t]
  \centering
  \includegraphics[width=0.92\columnwidth]{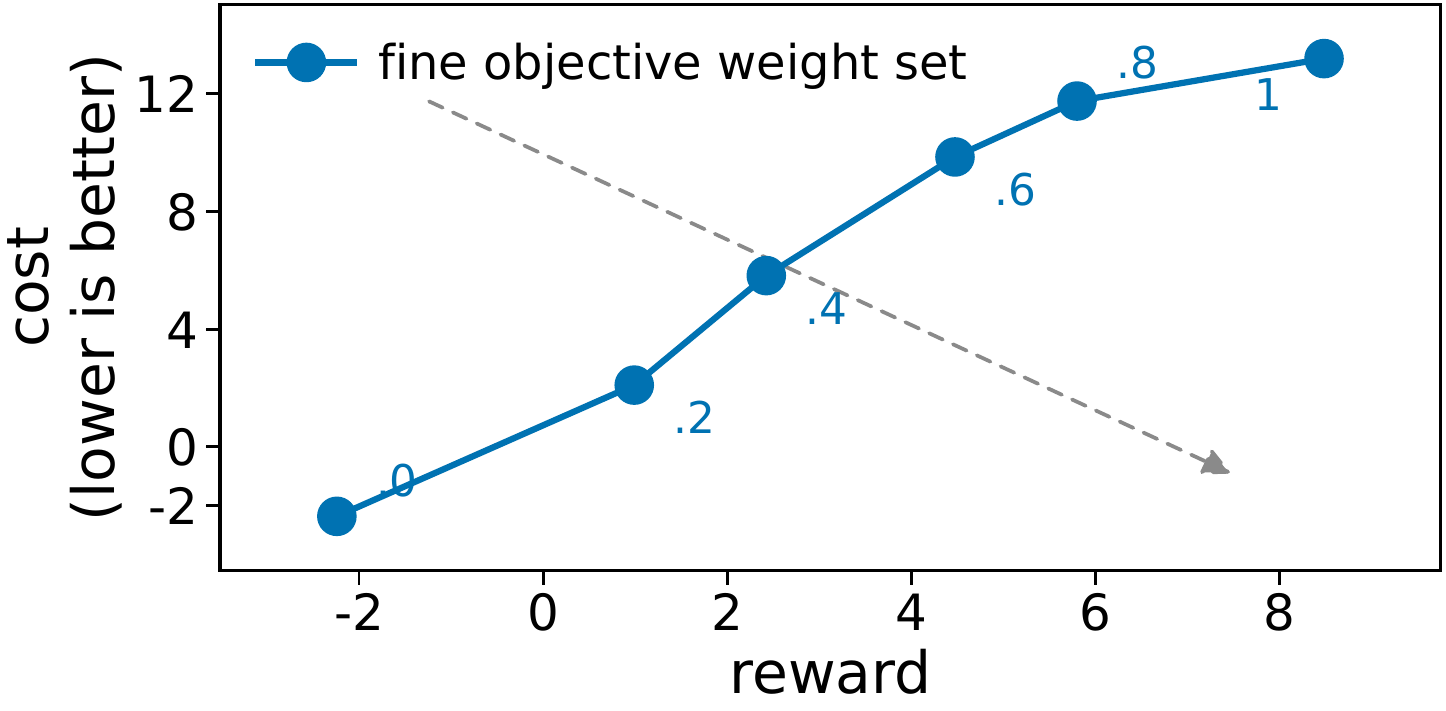}
  \caption{Reward vs.\ cost for the fine weight set on PKU-SafeRLHF (labels are objective weights; the $w{=}0$ point is the DPO margin adapter, since at $w{=}0$ all weight is on harmlessness). Higher reward is more helpful, lower cost is more harmless. Across these trained models mean reward increases and mean cost worsens with $w$ (a single-seed run), so no single model dominates and reaching a new target requires selecting or merging models.}
  \label{fig:front}
\end{figure}

\begin{figure*}[t]
  \centering
  \includegraphics[width=0.92\textwidth]{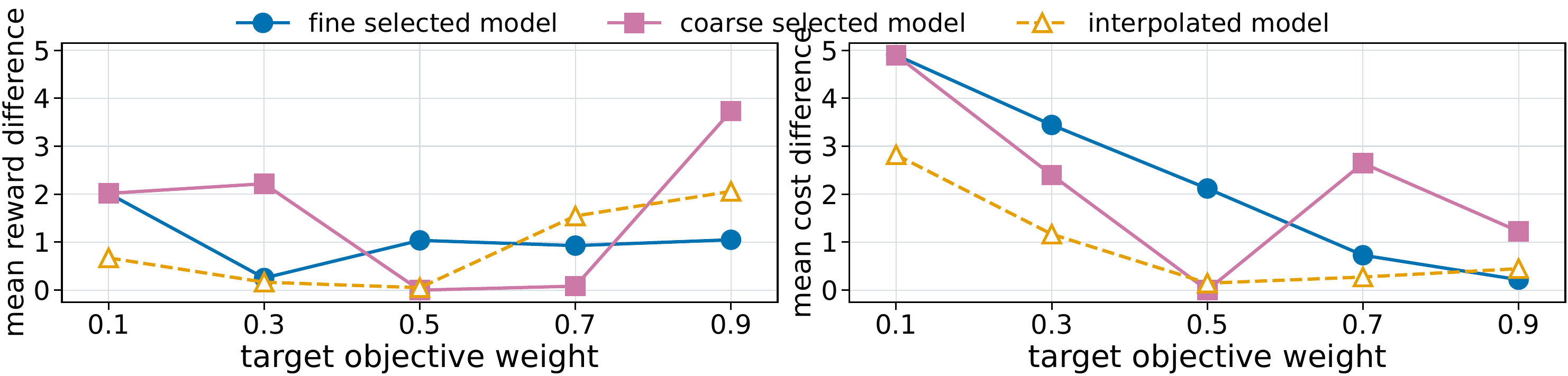}
  \caption{Absolute score difference from a directly trained target model at each target weight (lower is closer). Parameter merging (interpolated model, dashed) has the smallest combined gap but is not uniformly best: it is slightly worse than fine selection on mean reward, the selection methods spike at the $w^\star{=}0.1$ endpoint, and merging shows a large reward gap at $w^\star{=}0.9$. Merging is an interpolation heuristic, not an optimization procedure for the target weight.}
  \label{fig:scorediff}
\end{figure*}

Across the trained models, mean reward increases and mean harmlessness worsens as $w$ rises, an empirical trade-off frontier in this single-seed run (Figure~\ref{fig:front}), consistent with \citet{zhou-etal-2024-beyond}. No single model dominates, so reaching a new target is non-trivial. \Cref{tab:inference} and Figure~\ref{fig:scorediff} report the mean absolute difference from a model trained directly at the target. Parameter merging has the smallest average gap when reward and cost are considered together (combined mean $1.14$, vs.\ $1.69$ for fine selection and $2.40$ for coarse), although it is not consistently closest at individual target weights and is slightly worse than fine selection on mean reward gap ($1.11$ vs.\ $1.06$). Its $95\%$ confidence interval for the paired difference includes zero in $5$ of $10$ comparisons, vs.\ $2/10$ for fine selection: evidence of no detected difference, not of equivalence. Merging is an interpolation heuristic, not an optimization procedure for the target weight: its score trajectory did not reach the directly trained model at $w^\star{=}0.9$, producing a large reward gap, while the selection methods instead spike at the $w^\star{=}0.1$ endpoint. Nearest-model selection is limited by design: it picks on weight distance, not on scores, so the nearest weight is often not nearest in reward or
cost. \textbf{Both methods cover the continuum usefully, but neither reliably reproduces a directly trained model.}

\begin{table}[t]
\centering
\small
\setlength{\tabcolsep}{3pt}
\begin{tabular}{@{}lccc@{}}
\toprule
& \textbf{Fine} & \textbf{Coarse} & \textbf{Merge} \\
\midrule
Reward gap (mean$^\ast$)  & $\mathbf{1.06}$ & $2.01$ & $1.11$ \\
Cost gap (mean$^\ast$)    & $2.32$ & $2.79$ & $\mathbf{1.18}$ \\
CI incl.\ $0$             & $2/10$ & $1/8^\dagger$ & $\mathbf{5/10}$ \\
\bottomrule
\end{tabular}
\caption{Coverage study: mean absolute difference from direct training in Beaver reward and cost, and how many comparisons have a $95\%$ confidence interval for the paired difference that includes zero (no detected difference, not equivalence). Fine selection has the smallest mean reward gap; merging the smallest cost gap and combined gap. $^\ast$mean over target weights excluding $w^\star{=}0.5$, where the coarse set already contains the direct model.
$^\dagger$the coarse count is out of eight for the same reason: its two comparisons at $w^\star{=}0.5$ are zero by construction. Per-target values are in Appendix~\ref{sec:app-coverage}.}
\label{tab:inference}
\end{table}

\section{Discussion}
\label{sec:discussion}

\paragraph{Cheap measurements predict compatibility, but only for trustworthy data (RQ1).} The conflict rate and the reference-model correlation predict MODPO's behavior well on human-annotated HelpSteer: aligned objectives (correctness, coherence) improve together, and the most disagreeing pair (concision vs.\ helpfulness) trades off. They fail on AI-annotated UltraFeedback, but the failure is instructive. Rather than a clean objective incompatibility, the reward-model scores were strongly confounded by length and repetition, and once we controlled for those the results became substantially more consistent with the measurements.
We cannot conclude from this alone that the objectives are compatible; that would take an independent evaluator. For steerable pluralism this is a warning: whether two objectives \emph{look} compatible can depend heavily on how outputs are decoded and scored, so a system that decided which objectives to serve from a cheap measurement alone could draw the wrong line between compatible and conflicting objectives.

\paragraph{Recipe for RQ1.} To serve two objectives together, pick a pair with high correlation and low conflict, use the training-length limit that keeps the most pairs, and then \emph{check} whether the models generate very long or repetitive text and whether a different decoding setting changes the scores. To build a steerable trade-off, pick a pair with low correlation and high conflict, and use the margin sign to choose which direction of the objective to steer toward.

\paragraph{Coverage is cheap but approximate (RQ2).} A continuum of trade-offs can be covered by reusing a few trained models: parameter merging has the smallest combined gap from direct training and nearest-model selection is a simple fallback. Neither matches direct training: selection ignores scores, and merging is an interpolation heuristic rather than an optimization procedure for the target weight. This mirrors the trade-off in the wider merging literature \citep{jang2023personalizedsoups,rame2023rewardedsoups}: merging buys cheap coverage of the preference space at some cost in fidelity.

\paragraph{Recipe for RQ2.} Train a family of MODPO models across a range of weights. If it already covers the trade-offs you need, use nearest-model selection; otherwise consider filling the gaps by merging the two nearest models. When a target's exact behavior matters, train directly at that weight.

\paragraph{When is a trade-off the right frame?} Not every pair of objectives needs a steerable trade-off. Different stakeholders weight objectives differently, but when the objectives themselves reinforce each other (as correctness and helpfulness do here), a single well-trained model serves both, and offering a dial would be a false choice. A trade-off should be exposed only when the objectives genuinely compete for the same responses, which is what a high conflict rate, read together with the correlation and after controlling for length and repetition confounds, is meant to detect. This suggests a modest but useful role for the measurements in pluralistic systems: not to make the choice automatically, but to flag which pairs are candidates for steerable trade-offs and which are better served jointly, leaving the normative choice of \emph{which} trade-offs to expose to the people affected \citep{dönmez2026structuringspacesociotechnicalalignment}.

\paragraph{Takeaway.} An objective weight is a genuine steering dial, but it is not a promise: the same weight behaves differently depending on the data kept and the decoding used. Steerable pluralism with MODPO therefore needs its models characterized empirically by their generated length, repetition, and scored behavior, not just by their nominal weights.

\section{Conclusion}
\label{sec:conclusion}

We studied MODPO as a tool for steerable pluralistic alignment (one model family
that can be steered across a continuum of objective trade-offs to suit different
stakeholders) and asked when it can satisfy two objectives at once and how to
cover the continuum cheaply. Across seven objective pairs, two cheap pre-training
measurements predicted MODPO's behavior for human-annotated HelpSteer but not for
AI-annotated UltraFeedback, where response length and repetition strongly
confounded the reward-model scores; controlling for them made the results much
more consistent with the measurements, though an independent evaluator would be
needed to confirm compatibility. Whether two objectives can be served jointly is
therefore not something to read off a single number: it depends on the data, the
decoding, and the evaluator together. For covering new trade-offs
without retraining, nearest-model selection and parameter merging both help, but
neither reliably matches direct training. We hope the resulting recipes help
practitioners build models that serve diverse stakeholder preferences, and
measure honestly whether they do.

\section*{Limitations}
Our study has several limitations. (i) We steer only \emph{two} objectives at a time (helpfulness and a second objective); behavior with three or more as well as with different pairings may differ, and serving many stakeholders often means balancing many objectives at once. (ii) The coverage study reuses models within a fixed trained set and a fixed set of target weights, and parameter merging is tested only at the midpoint $\lambda=0.5$; other targets and merge positions are untested. (iii) All trained-model results use a \emph{single} seed, so we cannot quantify seed variation. (iv) All scores come from reward models, a proxy for human judgment. Every comparison we draw is within-evaluator: each MODPO model is read against the SFT reference trained on the same data and scored by the same model, with paired confidence intervals, so a constant evaluator bias shifts the score levels rather than the direction of a comparison. What this does not cover is absolute level and transfer, since ArmoRM is trained on data that includes HelpSteer and UltraFeedback and the Beaver models on the BeaverTails family that includes PKU-SafeRLHF. The one result that turns on the evaluator itself is the re-decoding comparison (\S\ref{sec:results-tradeoff}), where generation and scoring change together; we mark it as unsettled there. (v) We use one 7B base model per dataset, so results may transfer differently to other models or sizes. (vi) The conflict-split experiments change the training-set size and composition, so they are suggestive of a conflict-rate effect rather than causal. Promising next steps are a second (human or LLM-judge) evaluator to confirm objective scores independently of length and repetition, which matters most for UltraFeedback; a size-controlled study of the conflict rate; and a selection rule 2that uses reward and cost scores rather than weight distance alone.

\section*{Ethics Statement}
This work studies how to balance competing objectives in a single steerable model, using only publicly available datasets and models. Part of our motivation is pluralistic: making the trade-off between objectives (which different stakeholders would weight differently) \emph{explicit} and inspectable, rather than hiding it inside one aggregate reward. We note two risks. First, steerability cuts both ways: the same objective weight that balances helpfulness against harmlessness can be pushed toward helpfulness to produce less safe
outputs (\Cref{fig:front}), and our PKU-SafeRLHF setup includes adversarial prompts. Deciding \emph{which} trade-offs a deployed system should expose is a governance question beyond our scope. Second, reward-model scores carry the biases of their training data and, as we show, of response length and repetition, so they should not be read as ground-truth measures of any human value \citep{ghazaryan2026misalignedrewardsociallyundesirable}. We report these confounds openly and recommend human or independent evaluation before deployment.

\section*{Acknowledgments}
We acknowledge the support of the Ministerium für Wissenschaft, Forschung und Kunst BadenWürttemberg (MWK, Ministry of Science, Research and the Arts Baden-Württemberg under Az. 33-7533-9 19/54/5) in Künstliche Intelligenz \& Gesellschaft: Reflecting Intelligent Systems for Diversity, Demography and Democracy (IRIS3D) and the support by the Interchange Forum for Reflecting on Intelligent Systems (IRIS) at the University of Stuttgart. We would like to thank the Institute of Natural Language Processing (IMS) for providing computational support for this project.

\bibliography{bib/anthology-1, bib/custom}

\appendix
\section{Training and Evaluation Details}
\label{sec:appendix}

\paragraph{Data splits.} HelpSteer uses the official $95/5$ split
($48{,}888$ training / $2{,}474$ validation preference pairs); UltraFeedback has
no official validation split, so we use a random $90/10$ split with seed $0$
($345{,}414$ / $38{,}382$ pairs). PKU-SafeRLHF-10K is split into $9{,}000$
training and $1{,}000$ validation response pairs. Evaluation uses the first
$400$ unique validation prompts for the trade-off study and $700$ validation
response pairs ($665$ unique prompts) for the coverage study.

\paragraph{Training.} All stages use learning rate $10^{-4}$, KL coefficient
$\beta=0.1$, AdamW with a cosine schedule, weight decay $0.05$, and effective
batch size $8$ (one example, eight gradient-accumulation steps); the checkpoint
with the lowest validation loss is kept. Trade-off-study LoRA uses
rank/alpha/dropout $16/32/0.05$ on \texttt{q\_proj},\texttt{v\_proj} in bf16;
the coverage study uses $64/1/0$ on \texttt{q,k,v,o\_proj} in fp16. HelpSteer
trains $1{,}000$ steps per stage; UltraFeedback trains $300$ (SFT, margin) and
$400$ (MODPO); the coverage study trains $3$ epochs per stage with a
$1024$-token limit. We use split seed $0$, training seed $303$ (trade-off) or
$42$ (coverage), and generation seed $0$, with a single seed per model.

\paragraph{Generation.} Primary decoding is greedy with no repetition penalty or
$n$-gram blocking. The trade-off study generates up to $4096$ tokens
($1536$ prompt / $2560$ response); the coverage study up to $1024$. The
repetition-control setting adds repetition penalty $1.10$ and
\texttt{no\_repeat\_ngram\_size}$=4$.

\paragraph{Confidence intervals.} We use the percentile bootstrap over paired
score differences with $10{,}000$ resamples, seed $1000$, and $95\%$ intervals
\citep{efron1993bootstrap}, paired by validation prompt (trade-off) or response
pair (coverage).

\paragraph{Compute.} Training and evaluation used NVIDIA RTX A6000 GPUs for
approximately $1{,}252$ GPU-hours in total (${\sim}985$ for training,
${\sim}267$ for generation and scoring).

\section{Objective Compatibility Measurements}
\label{sec:app-counts}

\Cref{tab:app-counts} gives both pre-training measurements at both
training-length limits, together with the data they are computed from. The
512-token limit keeps roughly a third of the HelpSteer pairs and half of the
UltraFeedback pairs; the 2048-token limit keeps essentially all of them, which
is why we report it as primary.

\begin{table*}[t]
\centering
\small
\setlength{\tabcolsep}{4pt}
\begin{tabular}{@{}llrrrrrrr@{}}
\toprule
& & \textbf{Data} & \multicolumn{2}{c}{\textbf{SFT corr.}} & \multicolumn{2}{c}{\textbf{Conflict rate}} & & \\
\cmidrule(lr){4-5}\cmidrule(lr){6-7}
\textbf{Pair} & \textbf{Limit} & \textbf{corr.} & $512$ & $2048$ & $512$ & $2048$ & \textbf{Joint pairs} & \textbf{Retained} \\
\midrule
HS correctness & $512$ & $0.853$ & $0.982$ & $0.981$ & $0.022$ & $0.027$ & $7{,}033$ & $0.315$ \\
 & $2048$ & & & & & & $21{,}184$ & $0.996$ \\
HS coherence & $512$ & $0.635$ & $0.912$ & $0.911$ & $0.060$ & $0.067$ & $5{,}508$ & $0.317$ \\
 & $2048$ & & & & & & $16{,}340$ & $0.996$ \\
HS complexity & $512$ & $0.236$ & $0.178$ & $0.095$ & $0.148$ & $0.202$ & $4{,}168$ & $0.343$ \\
 & $2048$ & & & & & & $11{,}256$ & $0.996$ \\
HS verbosity ($+$) & $512$ & $0.256$ & $-0.068$ & $-0.162$ & $0.159$ & $0.226$ & $5{,}654$ & $0.335$ \\
 & $2048$ & & & & & & $15{,}663$ & $0.996$ \\
HS verbosity ($-$) & $512$ & $-0.256$ & $0.068$ & $0.162$ & $0.841$ & $0.774$ & $5{,}654$ & $0.335$ \\
 & $2048$ & & & & & & $15{,}663$ & $0.996$ \\
UF honesty & $512$ & $0.635$ & $0.870$ & $0.854$ & $0.186$ & $0.181$ & $93{,}218$ & $0.503$ \\
 & $2048$ & & & & & & $182{,}000$ & $0.995$ \\
UF instr.-following & $512$ & $0.698$ & $0.888$ & $0.879$ & $0.172$ & $0.164$ & $112{,}391$ & $0.513$ \\
 & $2048$ & & & & & & $216{,}157$ & $0.995$ \\
UF truthfulness & $512$ & $0.615$ & $0.832$ & $0.810$ & $0.193$ & $0.192$ & $87{,}402$ & $0.523$ \\
 & $2048$ & & & & & & $164{,}654$ & $0.994$ \\
\bottomrule
\end{tabular}
\caption{Pre-training measurements and the data behind them, for both training-length limits. \emph{Data corr.} is the Pearson correlation of the dataset's own scores; \emph{SFT corr.} the correlation of ArmoRM scores on reference-model generations (\S\ref{sec:measurements}). \emph{Joint pairs} counts response pairs that yield a preference pair for both objectives, and \emph{Retained} the share of valid response pairs the limit keeps. The 2048-token columns are the ones reported in \Cref{tab:measurements}.}
\label{tab:app-counts}
\end{table*}

\section{Per-Model Trade-off Scores}
\label{sec:app-scores}

\Cref{tab:app-scores} lists every score behind \Cref{fig:pairs},
\Cref{fig:margin}, and the \textbf{Both$\uparrow$?} column of
\Cref{tab:measurements}: all eight settings, both training-length limits,
and both decoding conditions. \Cref{tab:app-diag} gives the matching
length and repetition diagnostics.

\begin{table*}[t]
\centering
\small
\setlength{\tabcolsep}{4pt}
\begin{tabular}{@{}llccccccc@{}}
\toprule
\textbf{Pair} & \textbf{Limit} & \textbf{SFT} & $w{=}.1$ & $w{=}.2$ & $w{=}.4$ & $w{=}.6$ & $w{=}.8$ & $w{=}.9$ \\
\midrule
HS correctness & $512$ & $.628$/$.641$ & $.677$/$.681$ & $.680$/$.683$ & $.674$/$.677$ & $.621$/$.626$ & $.623$/$.627$ & $.614$/$.619$ \\
 & $2048$ & $.621$/$.636$ & $.702$/$.706$ & $.699$/$.703$ & $.683$/$.688$ & $.675$/$.680$ & $.668$/$.674$ & -- \\
HS coherence & $512$ & $.628$/$.737$ & $.684$/$.756$ & $.673$/$.749$ & $.651$/$.729$ & $.633$/$.713$ & $.628$/$.710$ & $.619$/$.698$ \\
 & $2048$ & $.621$/$.730$ & $.703$/$.770$ & $.692$/$.761$ & $.697$/$.763$ & $.676$/$.749$ & $.676$/$.748$ & -- \\
HS complexity & $512$ & $.628$/$.386$ & $.575$/$.439$ & $.606$/$.446$ & $.588$/$.443$ & $.609$/$.446$ & $.599$/$.436$ & $.604$/$.438$ \\
 & $2048$ & $.621$/$.382$ & $.626$/$.455$ & $.626$/$.451$ & $.646$/$.456$ & $.635$/$.452$ & $.653$/$.440$ & -- \\
HS verbosity ($+$) & $512$ & $.628$/$.427$ & $.631$/$.632$ & $.626$/$.630$ & $.605$/$.645$ & $.575$/$.653$ & $.603$/$.635$ & $.616$/$.636$ \\
 & $2048$ & $.621$/$.420$ & $.618$/$.625$ & $.606$/$.634$ & $.608$/$.641$ & $.602$/$.637$ & $.619$/$.622$ & $.661$/$.602$ \\
HS verbosity ($-$) & $512$ & $.628$/$.427$ & $.599$/$.334$ & $.620$/$.359$ & $.644$/$.384$ & $.660$/$.442$ & $.649$/$.597$ & $.624$/$.620$ \\
 & $2048$ & $.621$/$.420$ & $.573$/$.282$ & $.582$/$.289$ & $.628$/$.343$ & $.684$/$.428$ & $.696$/$.520$ & $.690$/$.547$ \\
UF honesty & $512$ & $.432$/$.518$ & $.438$/$.456$ & $.452$/$.471$ & $.450$/$.467$ & $.440$/$.453$ & $.441$/$.456$ & $.446$/$.455$ \\
 & $2048$ & $.430$/$.532$ & $.391$/$.394$ & $.388$/$.392$ & $.395$/$.398$ & $.410$/$.412$ & $.428$/$.424$ & -- \\
UF instr.-following & $512$ & $.432$/$.509$ & $.464$/$.504$ & $.470$/$.507$ & $.469$/$.497$ & $.459$/$.485$ & $.441$/$.467$ & $.436$/$.459$ \\
 & $2048$ & $.430$/$.515$ & $.375$/$.395$ & $.391$/$.409$ & $.418$/$.437$ & $.410$/$.432$ & -- & $.413$/$.428$ \\
UF truthfulness & $512$ & $.432$/$.581$ & $.461$/$.511$ & $.459$/$.512$ & $.462$/$.514$ & $.452$/$.498$ & $.459$/$.503$ & $.441$/$.484$ \\
 & $2048$ & $.430$/$.602$ & $.463$/$.507$ & $.451$/$.480$ & $.422$/$.444$ & $.421$/$.438$ & $.409$/$.419$ & -- \\
\midrule
\multicolumn{9}{@{}l}{\emph{2048-token models re-decoded with repetition control}} \\
HS correctness & $2048$ & $.600$/$.607$ & $.678$/$.673$ & $.677$/$.671$ & $.681$/$.675$ & $.678$/$.671$ & $.665$/$.661$ & -- \\
HS coherence & $2048$ & $.600$/$.722$ & $.667$/$.754$ & $.669$/$.754$ & $.668$/$.752$ & $.665$/$.750$ & $.668$/$.751$ & -- \\
HS complexity & $2048$ & $.600$/$.378$ & $.657$/$.463$ & $.659$/$.467$ & $.660$/$.461$ & $.664$/$.462$ & $.663$/$.444$ & -- \\
HS verbosity ($+$) & $2048$ & $.600$/$.386$ & $.652$/$.602$ & $.648$/$.605$ & $.653$/$.608$ & $.648$/$.593$ & $.656$/$.580$ & $.670$/$.558$ \\
HS verbosity ($-$) & $2048$ & $.600$/$.386$ & $.537$/$.274$ & $.547$/$.284$ & $.592$/$.330$ & $.645$/$.426$ & $.668$/$.505$ & $.674$/$.522$ \\
UF honesty & $2048$ & $.442$/$.535$ & $.544$/$.553$ & $.545$/$.553$ & $.548$/$.548$ & $.550$/$.551$ & $.554$/$.552$ & -- \\
UF instr.-following & $2048$ & $.442$/$.504$ & $.546$/$.536$ & $.548$/$.544$ & $.552$/$.546$ & $.556$/$.549$ & -- & $.552$/$.545$ \\
UF truthfulness & $2048$ & $.442$/$.595$ & $.543$/$.570$ & $.548$/$.574$ & $.551$/$.561$ & $.553$/$.554$ & $.550$/$.543$ & -- \\
\bottomrule
\end{tabular}
\caption{Mean ArmoRM scores over the same $400$ prompts, as helpfulness/second-objective. Leading zeros are dropped; \emph{--} marks a weight that was not trained. For HS verbosity ($-$) a lower second score is better. The lower block re-decodes the 2048-token models with repetition penalty $1.10$ and $4$-gram blocking; its aggregate is \Cref{tab:repcontrol}.}
\label{tab:app-scores}
\end{table*}

\begin{table*}[t]
\centering
\small
\setlength{\tabcolsep}{4pt}
\begin{tabular}{@{}llccccccc@{}}
\toprule
\textbf{Pair} & \textbf{Limit} & \textbf{SFT} & $w{=}.1$ & $w{=}.2$ & $w{=}.4$ & $w{=}.6$ & $w{=}.8$ & $w{=}.9$ \\
\midrule
HS correctness & $512$ & $174$/$.438$ & $321$/$.720$ & $338$/$.795$ & $374$/$.825$ & $588$/$.868$ & $550$/$.863$ & $597$/$.865$ \\
 & $2048$ & $202$/$.398$ & $257$/$.767$ & $263$/$.755$ & $315$/$.790$ & $348$/$.800$ & $317$/$.713$ & -- \\
HS coherence & $512$ & $174$/$.438$ & $313$/$.838$ & $376$/$.855$ & $438$/$.843$ & $513$/$.865$ & $520$/$.855$ & $560$/$.868$ \\
 & $2048$ & $202$/$.398$ & $234$/$.728$ & $269$/$.765$ & $272$/$.728$ & $292$/$.740$ & $299$/$.738$ & -- \\
HS complexity & $512$ & $174$/$.438$ & $788$/$.927$ & $628$/$.890$ & $736$/$.897$ & $622$/$.910$ & $647$/$.915$ & $626$/$.880$ \\
 & $2048$ & $202$/$.398$ & $544$/$.863$ & $551$/$.870$ & $477$/$.845$ & $505$/$.845$ & $371$/$.713$ & -- \\
HS verbosity ($+$) & $512$ & $174$/$.438$ & $478$/$.895$ & $492$/$.875$ & $589$/$.902$ & $716$/$.915$ & $634$/$.885$ & $592$/$.873$ \\
 & $2048$ & $202$/$.398$ & $528$/$.845$ & $578$/$.868$ & $610$/$.882$ & $645$/$.858$ & $597$/$.840$ & $429$/$.805$ \\
HS verbosity ($-$) & $512$ & $174$/$.438$ & $82$/$.233$ & $90$/$.263$ & $102$/$.345$ & $182$/$.450$ & $416$/$.790$ & $536$/$.875$ \\
 & $2048$ & $202$/$.398$ & $41$/$.138$ & $46$/$.145$ & $84$/$.278$ & $125$/$.445$ & $217$/$.640$ & $275$/$.703$ \\
UF honesty & $512$ & $329$/$.527$ & $727$/$.892$ & $687$/$.895$ & $677$/$.897$ & $695$/$.917$ & $683$/$.920$ & $670$/$.920$ \\
 & $2048$ & $234$/$.480$ & $897$/$.965$ & $889$/$.958$ & $862$/$.960$ & $808$/$.960$ & $786$/$.950$ & -- \\
UF instr.-following & $512$ & $329$/$.527$ & $590$/$.823$ & $584$/$.835$ & $607$/$.858$ & $642$/$.900$ & $687$/$.917$ & $706$/$.927$ \\
 & $2048$ & $234$/$.480$ & $972$/$.973$ & $922$/$.953$ & $824$/$.960$ & $828$/$.960$ & -- & $830$/$.970$ \\
UF truthfulness & $512$ & $329$/$.527$ & $640$/$.853$ & $629$/$.855$ & $623$/$.877$ & $656$/$.885$ & $618$/$.905$ & $686$/$.907$ \\
 & $2048$ & $234$/$.480$ & $599$/$.885$ & $683$/$.915$ & $804$/$.938$ & $787$/$.958$ & $820$/$.968$ & -- \\
\midrule
\multicolumn{9}{@{}l}{\emph{2048-token models re-decoded with repetition control}} \\
HS correctness & $2048$ & $88$/$.013$ & $164$/$.020$ & $161$/$.010$ & $161$/$.015$ & $164$/$.008$ & $149$/$.010$ & -- \\
HS coherence & $2048$ & $88$/$.013$ & $152$/$.023$ & $149$/$.015$ & $154$/$.020$ & $144$/$.010$ & $152$/$.013$ & -- \\
HS complexity & $2048$ & $88$/$.013$ & $210$/$.015$ & $204$/$.018$ & $196$/$.033$ & $197$/$.030$ & $156$/$.015$ & -- \\
HS verbosity ($+$) & $2048$ & $88$/$.013$ & $210$/$.030$ & $223$/$.030$ & $224$/$.048$ & $220$/$.040$ & $198$/$.025$ & $172$/$.028$ \\
HS verbosity ($-$) & $2048$ & $88$/$.013$ & $33$/$.003$ & $35$/$.000$ & $52$/$.003$ & $99$/$.003$ & $141$/$.008$ & $154$/$.013$ \\
UF honesty & $2048$ & $114$/$.080$ & $329$/$.305$ & $338$/$.313$ & $356$/$.348$ & $345$/$.320$ & $327$/$.298$ & -- \\
UF instr.-following & $2048$ & $114$/$.080$ & $397$/$.383$ & $383$/$.350$ & $354$/$.385$ & $346$/$.355$ & -- & $329$/$.338$ \\
UF truthfulness & $2048$ & $114$/$.080$ & $270$/$.233$ & $283$/$.265$ & $305$/$.315$ & $331$/$.310$ & $347$/$.353$ & -- \\
\bottomrule
\end{tabular}
\caption{Generation diagnostics for the same models, as mean words/share of responses with a repeated $4$-gram. Greedy decoding above, repetition control below. These are the numbers behind the length and repetition confound in \S\ref{sec:results-tradeoff}.}
\label{tab:app-diag}
\end{table*}

\section{Repetition Control, Per Pair}
\label{sec:app-repcontrol}

\Cref{tab:app-repcontrol} breaks \Cref{tab:repcontrol} out by
objective pair. The split is clean: no HelpSteer pair moves by more than
$0.036$ on either objective, no UltraFeedback pair by less than $0.103$, while
the length and repetition drops are comparable across the two datasets.

\begin{table*}[t]
\centering
\small
\setlength{\tabcolsep}{3pt}
\begin{tabular}{@{}lcccc@{}}
\toprule
& $\Delta$help & $\Delta$obj$_2$ & $\Delta$words & $\Delta$rep \\
\midrule
HS correctness & $-0.010$ & $-0.020$ & $-46.0\%$ & $-75.3$ \\
HS coherence & $-0.021$ & $-0.006$ & $-44.7\%$ & $-72.4$ \\
HS complexity & $+0.023$ & $+0.009$ & $-60.5\%$ & $-80.5$ \\
HS verbosity ($+$) & $+0.036$ & $-0.036$ & $-62.9\%$ & $-81.6$ \\
HS verbosity ($-$) & $-0.032$ & $+0.012$ & $-30.0\%$ & $-38.7$ \\
\midrule
UF honesty & $+0.145$ & $+0.147$ & $-59.9\%$ & $-64.2$ \\
UF instr.-following & $+0.149$ & $+0.124$ & $-58.7\%$ & $-60.1$ \\
UF truthfulness & $+0.116$ & $+0.103$ & $-58.2\%$ & $-63.8$ \\
\midrule
HelpSteer (mean) & $-0.001$ & $-0.008$ & $-48.8\%$ & $-69.7$ \\
UltraFeedback (mean) & $+0.137$ & $+0.125$ & $-58.9\%$ & $-62.7$ \\
\bottomrule
\end{tabular}
\caption{Per-pair version of \Cref{tab:repcontrol}: change in scores, mean words, and repeated-$4$-gram share (percentage points) when the 2048-token models are re-decoded with repetition control. Positive score changes are in the preferred direction.}
\label{tab:app-repcontrol}
\end{table*}

\section{Conflict-Split Results}
\label{sec:app-selection}

\Cref{tab:app-selection} gives the scores for the conflict-split
experiment of \S\ref{sec:selection}. The joint and non-conflict sets stay close
to the main results, as expected, since non-conflict pairs make up most of the
joint set; the conflict set is where the scores move.

\begin{table*}[t]
\centering
\small
\setlength{\tabcolsep}{4pt}
\begin{tabular}{@{}llccccc@{}}
\toprule
\textbf{Setting} & \textbf{Set} & \textbf{SFT} & $w{=}0.1$ & $w{=}0.2$ & $w{=}0.4$ & $w{=}0.8$ \\
\midrule
UF instr.-following, $512$ & joint & $.432$/$.509$ & $.445$/$.482$ & $.430$/$.463$ & $.456$/$.486$ & $.457$/$.483$ \\
 & non-conflict & $.432$/$.509$ & $.451$/$.475$ & $.458$/$.486$ & $.454$/$.480$ & $.446$/$.470$ \\
 & conflict & $.432$/$.509$ & $.412$/$.511$ & $.430$/$.522$ & $.427$/$.499$ & $.424$/$.453$ \\
\midrule
HS verbosity ($-$), $512$ & joint & $.628$/$.427$ & $.604$/$.348$ & $.604$/$.364$ & $.631$/$.398$ & $.647$/$.588$ \\
 & non-conflict & $.628$/$.427$ & $.595$/$.371$ & $.600$/$.393$ & $.621$/$.422$ & $.669$/$.562$ \\
 & conflict & $.628$/$.427$ & $.585$/$.653$ & $.617$/$.647$ & $.594$/$.650$ & $.614$/$.626$ \\
\midrule
HS verbosity ($-$), $2048$ & joint & $.621$/$.420$ & $.609$/$.317$ & $.608$/$.317$ & $.626$/$.345$ & $.693$/$.518$ \\
 & non-conflict & $.621$/$.420$ & $.581$/$.305$ & $.555$/$.277$ & $.613$/$.343$ & $.670$/$.462$ \\
 & conflict & $.621$/$.420$ & $.691$/$.419$ & $.694$/$.426$ & $.695$/$.441$ & $.701$/$.518$ \\
\bottomrule
\end{tabular}
\caption{Conflict-split results (\S\ref{sec:selection}): mean helpfulness/second-objective scores for margin reward models trained on the joint set, the non-conflict set (conflict rate $0$), and the conflict set (rate $1$). The split is computed in the dataset's score direction, so under the negative verbosity sign the ``conflict'' set is the one whose pairs agree with helpfulness.}
\label{tab:app-selection}
\end{table*}

\begin{table*}[t]
\centering
\small
\setlength{\tabcolsep}{4pt}
\begin{tabular}{@{}llccccc@{}}
\toprule
\textbf{Setting} & \textbf{Set} & \textbf{SFT} & $w{=}0.1$ & $w{=}0.2$ & $w{=}0.4$ & $w{=}0.8$ \\
\midrule
UF instr.-following, $512$ & joint & $329$/$.527$ & $714$/$.868$ & $738$/$.880$ & $673$/$.887$ & $643$/$.905$ \\
 & non-conflict & $329$/$.527$ & $676$/$.917$ & $685$/$.920$ & $667$/$.943$ & $685$/$.915$ \\
 & conflict & $329$/$.527$ & $212$/$.430$ & $270$/$.475$ & $471$/$.650$ & $732$/$.920$ \\
\midrule
HS verbosity ($-$), $512$ & joint & $174$/$.438$ & $100$/$.250$ & $119$/$.307$ & $138$/$.338$ & $435$/$.760$ \\
 & non-conflict & $174$/$.438$ & $155$/$.292$ & $186$/$.340$ & $184$/$.388$ & $280$/$.705$ \\
 & conflict & $174$/$.438$ & $756$/$.925$ & $606$/$.945$ & $707$/$.900$ & $595$/$.887$ \\
\midrule
HS verbosity ($-$), $2048$ & joint & $202$/$.398$ & $58$/$.217$ & $66$/$.212$ & $95$/$.263$ & $240$/$.598$ \\
 & non-conflict & $202$/$.398$ & $64$/$.217$ & $46$/$.155$ & $98$/$.292$ & $192$/$.490$ \\
 & conflict & $202$/$.398$ & $109$/$.390$ & $116$/$.417$ & $135$/$.445$ & $195$/$.605$ \\
\bottomrule
\end{tabular}
\caption{Length and repetition diagnostics for the conflict-split models of \Cref{tab:app-selection}, as mean words/share of responses with a repeated $4$-gram.}
\label{tab:app-selection-diag}
\end{table*}

\section{Coverage Study}
\label{sec:app-coverage}

\Cref{tab:app-coverage} gives the per-target gaps that
\Cref{tab:inference} averages, and \Cref{tab:app-coverage-abs} the
reward and cost scores they are computed from.

\begin{table}[t]
\centering
\footnotesize
\setlength{\tabcolsep}{4pt}
\begin{tabular}{@{}lcccccc@{}}
\toprule
& \multicolumn{3}{c}{\textbf{Reward gap}} & \multicolumn{3}{c}{\textbf{Cost gap}} \\
\cmidrule(lr){2-4}\cmidrule(lr){5-7}
$w^\star$ & Fine & Coarse & Merge & Fine & Coarse & Merge \\
\midrule
$0.1$ & $2.018$ & $2.018$ & $0.673$ & $4.900$ & $4.900$ & $2.822$ \\
$0.3$ & $0.253$ & $2.219$ & $0.166$ & $3.444$ & $2.400$ & $1.167$ \\
$0.5$ & $1.038$ & -- & $0.051$ & $2.119$ & -- & $0.146$ \\
$0.7$ & $0.928$ & $0.084$ & $1.546$ & $0.728$ & $2.645$ & $0.273$ \\
$0.9$ & $1.050$ & $3.732$ & $2.059$ & $0.218$ & $1.226$ & $0.448$ \\
\midrule
mean$^\ast$ & $1.062$ & $2.013$ & $1.111$ & $2.323$ & $2.793$ & $1.178$ \\
\bottomrule
\end{tabular}
\caption{Per-target absolute difference from a model trained directly at $w^\star$, the values plotted in \Cref{fig:scorediff} and averaged in \Cref{tab:inference}. \emph{--} marks the targets where coarse selection returns the direct model itself. $^\ast$mean excluding $w^\star{=}0.5$.}
\label{tab:app-coverage}
\end{table}

\begin{table}[t]
\centering
\footnotesize
\setlength{\tabcolsep}{1.7pt}
\begin{tabular}{@{}lcccccccc@{}}
\toprule
& \multicolumn{2}{c}{\textbf{Direct}} & \multicolumn{2}{c}{\textbf{Lower}} & \multicolumn{2}{c}{\textbf{Merge}} & \multicolumn{2}{c}{\textbf{Upper}} \\
\cmidrule(lr){2-3}\cmidrule(lr){4-5}\cmidrule(lr){6-7}\cmidrule(lr){8-9}
$w^\star$ & rew. & cost & rew. & cost & rew. & cost & rew. & cost \\
\midrule
$0.1$ & $-0.22$ & $2.53$ & $-2.24$ & $-2.37$ & $-0.90$ & $-0.29$ & $0.99$ & $2.10$ \\
$0.3$ & $1.24$ & $5.54$ & $0.99$ & $2.10$ & $1.41$ & $4.37$ & $2.42$ & $5.82$ \\
$0.5$ & $3.46$ & $7.94$ & $2.42$ & $5.82$ & $3.51$ & $8.09$ & $4.47$ & $9.86$ \\
$0.7$ & $3.54$ & $10.58$ & $4.47$ & $9.86$ & $5.09$ & $10.86$ & $5.80$ & $11.76$ \\
$0.9$ & $4.75$ & $11.97$ & $5.80$ & $11.76$ & $6.81$ & $12.42$ & $8.48$ & $13.20$ \\
\bottomrule
\end{tabular}
\caption{Mean Beaver reward and cost over the $700$ coverage prompts for the directly trained model, the two neighbouring trained models, and their merge. Every merged model lands between its neighbours, which is why merging cannot reach a direct target outside that interval (reward at $w^\star{=}0.7,0.9$; cost at $w^\star{=}0.1$).}
\label{tab:app-coverage-abs}
\end{table}

\end{document}